# IsraParlTweet: The Israeli Parliamentary and Twitter Resource

Guy Mor-Lan[1], Effi Levi[2], Tamir Sheafer[1, 3], Shaul R. Shenhav[1]

[1]Department of Political Science, The Hebrew University of Jerusalem
[2]Institute of Computer Science, The Hebrew University of Jerusalem
[3]Department of Communication and Journalism, The Hebrew University of Jerusalem

{guy.mor, tamir.sheafer, shaul.shenhav}@mail.huji.ac.il, efle@cs.huji.ac.il

**Abstract**

We introduce IsraParlTweet, a new linked corpus of Hebrew-language parliamentary discussions from the Knesset (Israeli Parliament) between the years 1992-2023 and Twitter posts made by Members of the Knesset between the years 2008-2023, containing a total of 294.5 million Hebrew tokens. In addition to raw text, the corpus contains comprehensive metadata on speakers and Knesset sessions as well as several linguistic annotations. As a result, IsraParlTweet can be used to conduct a wide variety of quantitative and qualitative analyses and provide valuable insights into political discourse in Israel.

## 1. Introduction

Language resources based on parliamentary records are gaining increasing attention from NLP and social science scholars (Eide, 2020; Erjavec et al., 2021; Escribano et al., 2022; Abrami et al., 2022). Derived from the French word *parler* ("to talk"), parliaments have long been a main venue for open and public debates. The systematic documentation of such a major political arena for public deliberation provides a wealth of research opportunities for scholars from all areas of academic research. In parliaments that speak low-resource languages (such as Hebrew), these resources are of particular importance, since they provide a rare opportunity to gather spoken language corpora (Agić and Vulić, 2019).

Parliaments have never been the only venue for public political communication. For example, a long history of political communication research that goes back to the classical work of Laswell (1948) paid attention to the use of mass media by politicians (Barberá and Zeitzoff, 2018). The use of social media for political communication has received an increasing amount of attention in recent years (Barberá and Zeitzoff, 2018). Social media has been studied as a communication platform for politicians, providing them with an alternative outlet for sending messages and setting political agenda (Gilardi et al., 2022). Twitter[1], in particular, has been shown to offer such opportunities for members of parliament, providing a way to circumvent constraints present in other communication outlets, as well as providing a way to 'amplify' party messages (Silva and Proksch, 2022).

With this in mind, we introduce IsraParlTweet, a new digital language resource that documents Knesset (Israeli parliament) discussions over the years 1992-2023, as well as Twitter posts by Knesset members in the years 2008-2023.[2] The text is augmented by rich meta-data such as personal details, political roles & affiliations, and Twitter statistics, allowing a diverse array of possible statistical analyses. In addition to exploring each of the communication outlets separately, the significant overlap in time periods allows fascinating explorations and analyses of the inter-connections and interactions of members' expressions in the Knesset vs. their expressions in Twitter. Consisting of a total of 294.5 million tokens in Hebrew, IsraParlTweet is also one of the largest catalogued language resources in a language that has very few such resources available for researchers.

The remainder of this paper is organized as follows: Section 2 provides a short description of the Knesset and its relevant aspects covered by the corpus. Section 3 describes the process of creating the corpus. Section 4 contains a detailed description of the corpus. Section 5 provides details on several statistical analyses performed on the data as well as a discussion of the results. Finally, Section 6 provides a conclusion to the paper.

## 2. The Knesset

Israel is a parliamentary democracy. The Israeli parliament, "The Knesset", is a unicameral legislature with 120 members of Knesset (MKs) elected in general elections (Galnoor and Blander, 2018; Hazan, 2022). The Knesset elections are supposed to take place every four years, but early elections are very common. The Israeli electoral system is characterized by three major features: a proportional allocation formula; the use of a single nationwide district for seat allocation, and a closed-party list system (Hazan, 2022). The Israeli political system is regarded as a prototype of a PR (proportional representation) system (Hazan et al., 2017) and its multi-party system is characterized by intense ideological competition (Oshri et al., 2021). Party lists represented in the Knesset range between 8 (the 2020 elections) and 15 (the 1999 elections). The number of parties represented in the Knesset can be even higher when lists have more than one party or when parties split during a Knesset term (Kenig and Rahat, 2023). As a result, the Knesset reflects the heterogeneity of Israeli society. There are usually three Knesset plenary meetings each week: on Mondays, Tuesdays, and Wednesdays. Knesset meetings are conducted in Hebrew, but Members can speak in other languages and the protocol will include translations into Hebrew of such speeches[3]. Knesset meetings are regularly attended by government ministers and the prime minister, who is by law a member of the Knesset.

---

[1]Changed to "X" on July 2023.

[2]https://github.com/guymorlan/IsraParlTweet

Knesset debates cover a variety of topics, including security, economics, and social issues. These debates also cover a variety of types of deliberation, among them are legislation, motions for the agenda, parliamentary questions for government Ministers, and motions of no-confidence in the government[4].

## 3. Corpus Creation

### 3.1. Knesset Protocols

#### 3.1.1. Preprocessing

The Knesset Plenary Records ("Divrei Haknesset") is a verbatim record of the debates held in the Knesset Plenary Chamber. The Knesset regulations direct that the protocol includes discussions above the stand or in designated places in the plenary hall, as well as interjections. We processed the digital versions of Knesset plenary protocols publicly available as DOC and DOCX files.[5]

Speeches from the Knesset protocols were extracted as far as the 13th Knesset, sworn in on July 13th, 1992. Extraction started at that point in time since DOC/DOCX from prior Knessets were digitized via OCR and could not be parsed reliably given the amount of OCR errors. DOCX files were parsed using the `python-docx` python module[6], while DOC files were converted to DOCX with Microsoft Word, using the `win32com` python client.

An iterative pipeline has been developed to identify and extract topics, speaker names and speech texts utilizing style tags, stylistic cues present in the documents and regular expressions. This procedure has been tailored to account for changes over time in Knesset protocol formats, resulting in two separate pipelines for processing the data.

**Pipeline A**. This pipeline has been designed for newer formats of the protocol containing style tags produced by various document-generating systems throughout the years. These style tags were used to identify different parts of each protocol. However, the tags and their names vary considerably throughout the years. In order to capture these changes we have conducted an iterative process based on the following steps: (1) Extracting information based on an initial list of style tags, (2) Human validation of a sample of raw protocol text and extracted information from different Knessets to identify misidentification, and (3) updating list of style tags and regular expressions. Steps 1-3 were repeated until no faults were found and extracted information matched raw protocols.

**Pipeline B**. This pipeline was developed for older formats in which no style tags were used. In their absence, extraction in these cases was based on stylistic cues. As with style tags in the newer format, stylistic cues have changed throughout the years in these older files. In order to capture these changes we have conducted an iterative process based on the following steps: (1) Extracting information based on an initial list of stylistic cues present in the documents and regular expressions, (2) Human validation of sample of text from different Knessets to identify misidentification, and (3) updating processing of stylistic cues and regular expressions. Steps 1-3 were repeated until no faults were found and extracted information matched raw protocols.

#### 3.1.2. Metadata

Speaker names extracted from the protocols were matched to a metadata file containing longitudinal data on MKs, including time of service as MKs in each Knesset and each party (in case of changes in party affiliation), full name, gender, date and birth and place of birth. Names of speakers were matched to canonical names from the metadata file via exact matching when possible, or otherwise using fuzzy string matching with human validation.

Efforts were made so that the extracted text and metadata match the plenary protocols as closely as possible. There was no attempt to find and fix mistakes or issues with the plenary documents themselves, except for a handful of systematic issues (e.g., the '=' character appearing instead of '-' in older protocol files).

---

[3] Knesset regulations, "Takanon ha-knesset, October 16, 2023, chapter 5 38(b). https://main.knesset.gov.il/Activity/Documents/RulesOfProcedure.pdf

[4] The Plenum at Work. https://main.knesset.gov

[5] https://main.knesset.gov.il/Activity/plenum/Pages/Sessions.aspx

[6] https://github.com/python-openxml/python-docx

## 3.2. Twitter

In order to facilitate Twitter data collection, we first recorded the Twitter handles of all Israeli MKs serving during the 20th to the 25th Knessets. We applied for, and were granted, an Academic Research Access to the Twitter API v2[7], and utilized it to download all the posts made by these accounts, from the beginning of their activity on Twitter (as far back as 2008). The extraction was carried out using the `tweepy` python module[8]. Twitter post collection concluded on 27.3.2023, and the corpus reflects the availability of posts on that date. Twitter handles were manually associated with identifiers from the metadata file.

## 4. The Corpus

The corpus is divided into four main sections: Knesset Sessions, Twitter Posts, Office Sessions and Linguistic Analyses. Here, we provide a detailed description for each of the sections.

### 4.1. Knesset Sessions

This section contains the utterances of the MKs on the Knesset floor, in the order in which they appeared in the consecutive plenary protocol files. The utterances vary in length and may contain anything from a few words to a complete speech. Interruptions and interjections are preserved as they appear in the protocols. In total, this section contains approximately 4.5M individual utterances. The data is organized in CSV format, where each row represents a single utterance, and contains the following fields:

- **text**. The text of the utterance.
- **uuid**. A unique text identifier used for associating the text with separately provided morphological analysis.
- **knesset**. Knesset term.
- **session_number**. Session number in current Knesset term.
- **date**. Date of session.
- **person_id**. Numeric identifier for speaker. Numeric identifiers are only assigned to MKs. 3% of speakers lack an identifier in cases of non-MK politicians (e.g. president, non-MK ministers), administrative Knesset workers or guests, or if the matching MK could not be determined. Speakers are assigned an identifier if they were MKs in the time period of the corpus, even if they are not MKs at the time the utterance is made (e.g. presidents that were previously MKs).
- **canonical_name**. The canonical name (first name and surname) of the speaker. Only present for MKs for which an identifier can be determined.
- **name**. The name of the speaker as extracted from the protocol.
- **chair**. Indicator for whether or not the speaker was the chair of session.
- **topic**. Topic of discussion or agenda item.
- **topic_extra**. Additional information on the topic (e.g. subtitle, legislation proposal number).
- **qa**. Indicator for whether or not the utterance is part of a Questions and Answers session.
- **query**. The written query to which utterance is a an oral response.
- **only_read**. Indicator for whether or not the utterance was a Q&A response that was read and not delivered by the answerer orally.

Table 1 displays statistics for the Knesset Sessions section of the corpus, organized by Knesset terms.

### 4.2. Twitter Posts

This section contains the tweets posted by the MKs, in chronological order. In total, it contains approximately 821K tweets. The data is organized in a CSV format, where each row represents a single tweet and contains the following fields:

- **text**. The text of the tweet.
- **uuid**. A unique text identifier used for associating the text with separately provided morphological analysis.
- **tweet_id**. Twitter's unique tweet identifier.
- **date**. Date of the tweet.
- **knesset**. The Knesset term corresponding to the date of the tweet.
- **person_id**. Numeric identifier for the tweet poster. All rows have an identifier since only posts by MKs were collected. However, note that the poster was not necessarily serving as an MK at the time of posting.
- **user_id**. Twitter user ID number.
- **username**. Twitter handle name.

---

[7] Discontinued on April 2023.
[8] https://github.com/tweepy/tweepy

| Knesset | Start Date | End Date | Sessions | Texts | Unique Speakers | Total Tokens |
|---|---|---|---|---|---|---|
| 13 | 13.6.1992 | 17.6.1996 | 436 | 1.19M | 135 | 76.85M |
| 14 | 17.6.1996 | 7.6.1999 | 287 | 795.15K | 139 | 45.13M |
| 15 | 7.6.1999 | 17.2.2003 | 356 | 939.76K | 149 | 53.98M |
| 16 | 17.2.2003 | 17.4.2006 | 311 | 264.47K | 143 | 17.92M |
| 17 | 17.4.2006 | 24.2.2009 | 277 | 140.01K | 135 | 11.03M |
| 18 | 24.2.2009 | 5.2.2013 | 386 | 300.36K | 140 | 18.31M |
| 19 | 5.2.2013 | 31.3.2015 | 193 | 129.89K | 137 | 8.28M |
| 20 | 31.3.2015 | 30.4.2019 | 372 | 325.97K | 149 | 17.93M |
| 21 | 30.4.2019 | 3.10.2019 | 28 | 7.25K | 117 | 512.29K |
| 22 | 3.10.2019 | 16.3.2020 | 28 | 9.41K | 119 | 635.30K |
| 23 | 16.3.2020 | 6.4.2021 | 138 | 73.97K | 146 | 4.87M |
| 24 | 6.4.2021 | 15.11.2022 | 155 | 212.58K | 156 | 10.10M |
| 25 | 15.11.2022 | 16.8.2023* | 102 | 103.39K | 147 | 5.52M |
| **Total** | – | – | 3,096 | 4.49M | 637 | 271.07M |

\* Data collection concluded on this date.

Table 1: Knesset Data

| Knesset | Full Coverage | Start Date | End Date | Tweets | Unique Speakers | Total Tokens |
|---|---|---|---|---|---|---|
| 17 | × | 7.7.2008* | 24.2.2009 | 409.00 | 3 | 6.41K |
| 18 | × | 24.2.2009 | 5.2.2013 | 28.07K | 50 | 562.44K |
| 19 | × | 5.2.2013 | 31.3.2015 | 36.15K | 76 | 716.54K |
| 20 | ✓ | 31.3.2015 | 30.4.2019 | 289.88K | 183 | 7.27M |
| 21 | ✓ | 30.4.2019 | 3.10.2019 | 41.15K | 167 | 1.28M |
| 22 | ✓ | 3.10.2019 | 16.3.2020 | 44.58K | 178 | 1.38M |
| 23 | ✓ | 16.3.2020 | 6.4.2021 | 137.71K | 191 | 4.21M |
| 24 | ✓ | 6.4.2021 | 15.11.2022 | 196.16K | 193 | 6.41M |
| 25 | ✓ | 15.11.2022 | 27.3.2023** | 47.64K | 182 | 1.59M |
| **Total** | – | – | – | 821.34K | 235 | 23.43M |

\* Data collection started on this date.
\*\* Data collection concluded on this date.

Table 2: Twitter Data

- **name**. The canonical name (first name + surname) of the poster.

- **likes**. Number of likes received at collection time.

- **retweets**. Number of retweets at collection time.

- **replies**. Number of replies at collection time.

- **quotes**. Number of quotes at collection time.

Table 2 displays statistics for the Twitter Data section of the corpus, organized by Knesset terms. Terms 20-25 are fully covered in that all Twitter posts for all MKs serving in these terms were collected. Terms 17-19 are partially covered as only posts made by MKs serving in the 20-25 Knesset terms were collected.

### 4.3. Office Sessions

This section contains metadata describing the office sessions of the MKs. An office session is a period of time in which a person served as an MK under a given party or faction. The data is organized in CSV format, where each row, representing a single office session, contains the following fields:

- **start_date**. Start date of office session.

- **end_date**. End date of office session.

- **knesset**. Relevant Knesset term.

- **person_id**. A unique personal id used for matching with Knesset Session utterances and Twitter Posts.

- **first_name**. MK's first name.

- **surname**. MK's surname.

- **gender**. MK's gender.
- **faction**. Name of faction under which the MK served.
- **faction_id**. Unique identifier for faction.
- **party_name**. Unified party name under which the MK served.
- **dob**. MK's date of birth.
- **cob**. MK's country of birth.
- **yod**. MK's year of death.
- **yoi**. MK's year of immigration (Aliyah) to Israel.
- **city**. MK's city of residence.
- **languages**. MK's spoken languages – a comma separated string.

| POS Tag | Knesset (%) | Twitter (%) |
|---|---|---|
| NOUN | 21.65 | 21.20 |
| PUNCT | 14.27 | 11.00 |
| ADP | 13.72 | 15.11 |
| DET | 10.36 | 10.40 |
| VERB | 9.72 | 9.31 |
| PRON | 7.11 | 4.78 |
| ADV | 5.69 | 5.38 |
| SCONJ | 3.73 | 2.99 |
| PROPN | 3.59 | 6.71 |
| ADJ | 3.57 | 4.27 |
| CCONJ | 3.36 | 3.24 |
| AUX | 1.84 | 1.39 |
| NUM | 1.38 | 4.22 |
| INTJ | 0.01 | 0.003 |
| X | 0.0008 | 0.0025 |

Table 3: Distribution of POS Tags in Knesset and Twitter Corpora

### 4.4. Linguistic Analyses

This section contains additional analyses performed on the texts in the the Knesset Sessions and Twitter Data sections. The texts were processed using the Stanza morphological analysis pipeline (Qi et al., 2020). Sentiment labels were added using the HeBERT Hebrew sentiment model (Chriqui and Yahav, 2022). The data is organized in JSON format, where the keys are the 'uuid' identifiers for individual texts in the Knesset Sessions and Twitter Posts sections. The values include:

- **sentences**. The text segmented into a list sentences.
- **tokens**. A list of tokens for each sentence. Each token is a dictionary containing Stanza's morphological outputs:
  - *id*. Token id.
  - *text*. Token text.
  - *lemma*. Token lemma.
  - *upos*. Universal part-of-speech tag.
  - *xpos*. Treebank-specific part-of-speech tag.
  - *feats*. A list of grammatical features, e.g. gender, Binyan, Number, Person, Tense and Voice.
  - *misc*. A list of starting character position and ending character position for the tokens in the text.
- **sentiment** A list of sentiment labels for each sentence (negative/positive/neutral).

It is important to note that while the corpus is overwhelmingly in the Hebrew language, occasional texts may partially or fully consist of tokens in English, Arabic, Amharic or other languages.

## 5. Data Analysis

The fact that the textual data was taken from two different sources – Knesset sessions and Twitter posts – as well as its augmentation with rich and diverse metadata, allows for a wide variety of possible interesting analyses. In this section, we describe several such analyses and discuss their results.

### 5.1. Platform (Knesset vs. Twitter)

Table 3 shows the distribution of part-of-speech tags in the Knesset and Twitter sections. Although there are no drastic differences, it is interesting to note those which do exist between the sources. For example, Knesset utterances contain more pronouns, pointing to a more prominent use of first-person expressions. Twitter posts, on the other hand, contain more proper nouns, indicating a more person-focused discourse.

In order to approximate the lexical diversity across the two different platforms, we use the common Type/Token Ratio (TTR) measure (Chotlos, 1944), calculated as $TTR = \frac{T_{types}}{T_{tokens}}$, where $T_{types}$ is the number of unique tokens and $T_{tokens}$ is the total number of tokens. A higher TTR is considered to indicate higher lexical diversity. Figure 3 shows the histograms of the TTR (per MK) for the Knesset and Twitter sections. The results indicate that Twitter posts are generally more lexically diverse ($TTR = 0.22$) than Knesset sessions ($TTR = 0.13$).

Figure 4 shows the histogram of speaker sentiment scores in Knesset floor speeches and Twitter posts. This histogram was calculated for the set of speakers for which both Knesset and Twitter data is available. For each MK, we aggregated the sentence-level sentiment labels to calculate a sentiment score $S = \frac{p-n}{p+n}$, where $p$ is the count of positive sentences and $n$ the count of negative sentences. As can be seen, while speech on both platforms is overwhelmingly negative, Twitter speech ($S = -0.32$) is more positive than speech on the Knesset floor ($S = -0.54$).

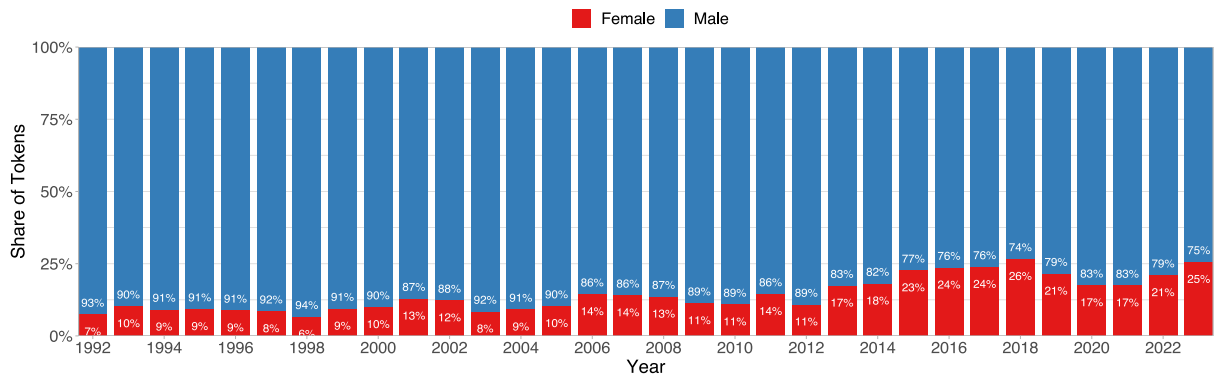

Figure 1: Token Percentage – Knesset

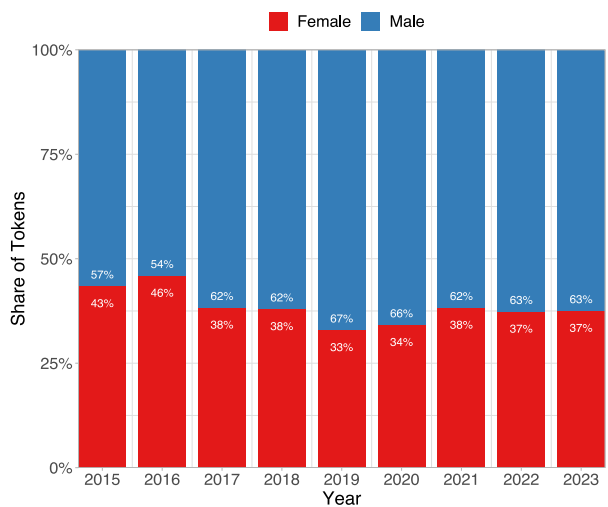

Figure 2: Token Percentage – Twitter

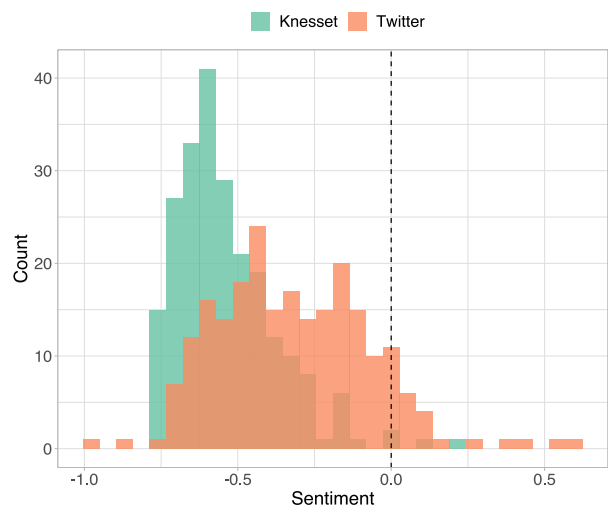

Figure 4: Sentiment – Twitter vs Knesset

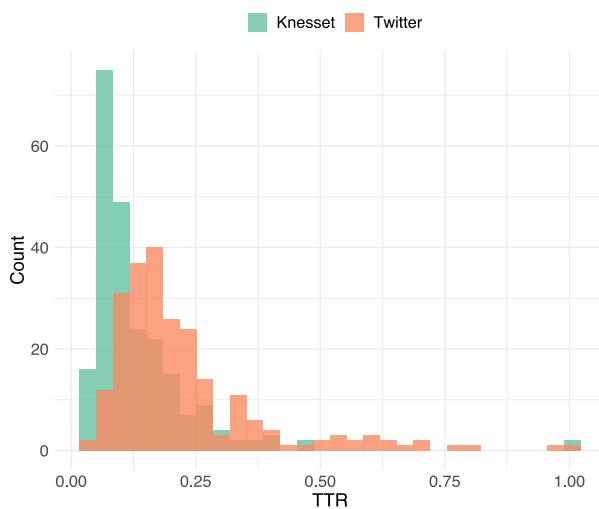

Figure 3: TTR – Knesset vs Twitter

### 5.2. Gender

Figure 1 displays the yearly percentage of tokens produced by women and by men in the Knesset. It is evident that while women's speech as measured in tokens remains a minority, there is a significant increase in the share of tokens spoken by women across time.

Similarly, Figure 2 shows the the yearly percentage of tokens produced by women and by men on Twitter. The results show that women have generated a significantly larger share of Twitter tokens (compared to Knesset utterances), albeit less than 50%. This share has actually decreased from 2016 onwards.

Figure 5 displays the yearly TTR by gender for the Knesset section, and Figure 7 displays the yearly TTR by gender for the Twitter section. For each platform, the TTR values were calculated on the aggregated tokens for all the MKs of the same gender. The results show that women's speech has a consistently higher lexical diversity both in the Knesset and in Twitter posts. Nevertheless, men and women's lexical diversity follow a common trend over time on both platforms.

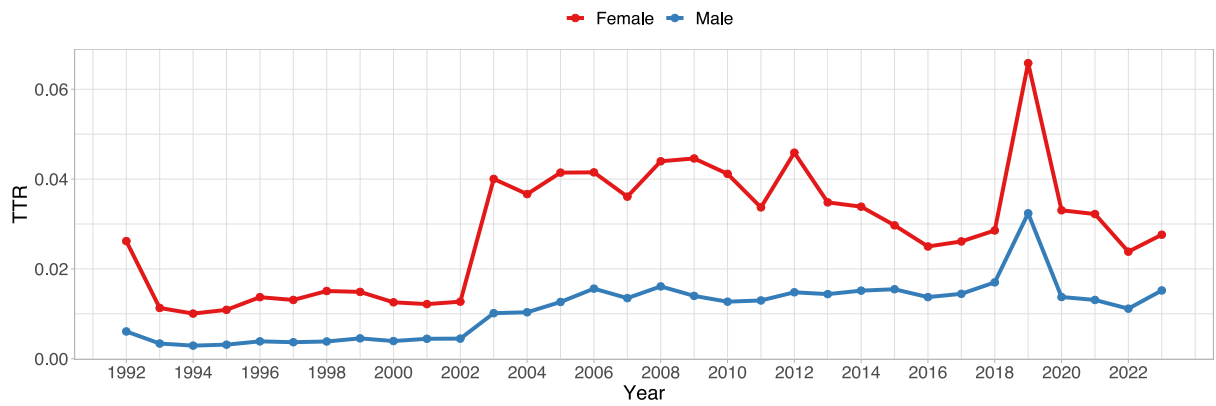

Figure 5: TTR – Knesset

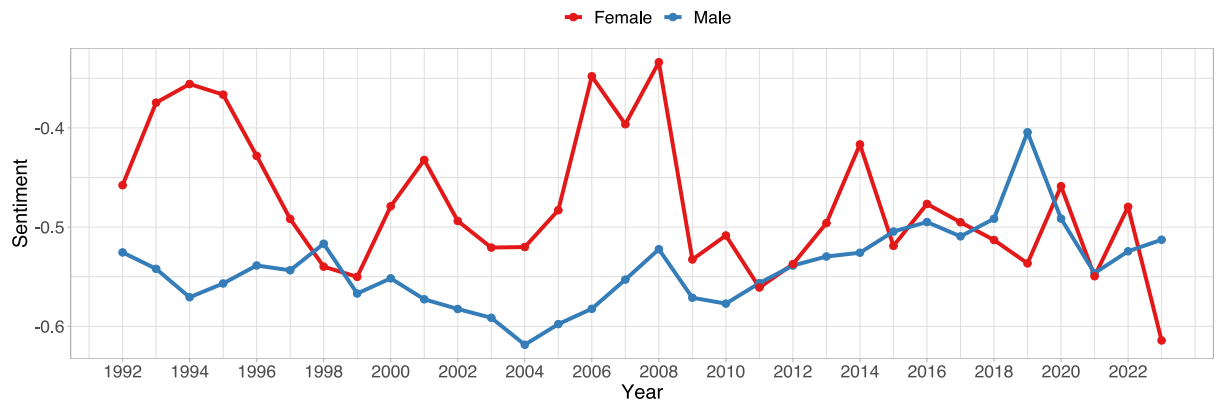

Figure 6: Sentiment – Knesset

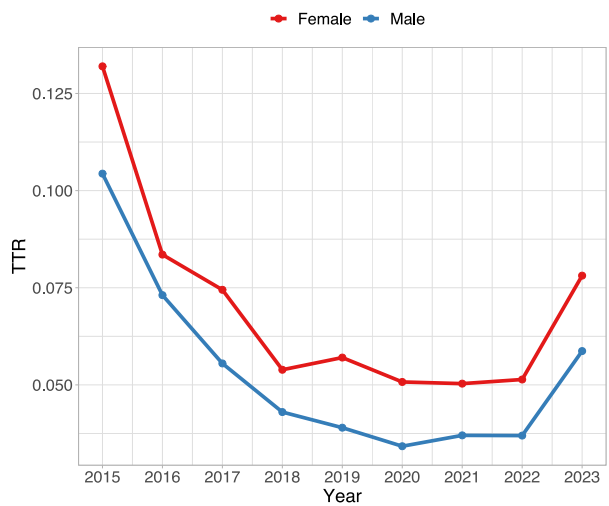

Figure 7: TTR – Twitter

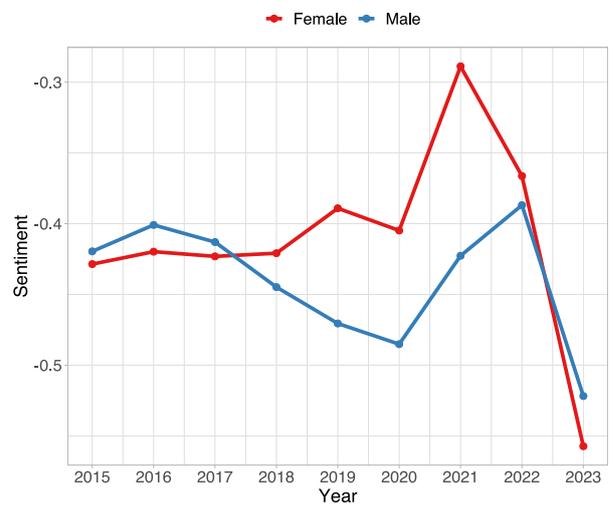

Figure 8: Sentiment – Twitter

Figure 6 shows the yearly mean sentiment score by gender for the Knesset section, and Figure 8 displays the yearly mean sentiment score by gender for the Twitter section. While both women and men's speech on both platforms is generally neg-ative, women's speech has a more positive sentiment overall. In Knesset utterances, the first 20 years show a significant difference in sentiment, whereas the latter 10 years show convergence to a similar sentiment. During these same 10 years, however, a significant difference remains evident between men and women in Twitter posts.

Finally, we examine which words most distinctively characterize female and male MKs on each of the platforms. We do so by following Monroe et al. (2008) in calculating the *weighted log odds* for each token in the text. Table 4 shows the top 20 distinctive lemmatized tokens for women and men in Knesset utterances, and Table 5 reports the same for Twitter posts. Both tables show the leading distinctive tokens after removal of non-informative tokens. In order to accommodate non-Hebrew readers, each lemma has also been translated into English.

We can make several interesting observations based on these two tables. Observing the first table, female Knesset discourse seems to be centered around family, gender and social issues (domestic violence, environmental problems), whereas male Knesset discourse is more focused on issues of security, economics and the Jewish religion (the latter is probably due to the fact that religious orthodox factions consist solely of male MKs). Female utterance vocabulary also exhibits more attention to emotions compared to their male counterparts.

The second table provides an interesting comparison as well. On Twitter, both genders use terms that seem to elicit emotional responses. However, female MKs tend to use terms that are more associated with individual and family-related environments, as well as words and emojis that describe emotional states. Male MKs, on the other hand, tend to appeal more to the "collective" and partisan-political environments, using terms related to security and national identity. Another prominent difference is in the use of proper nouns – male posters tend to utilize names much more than female posters (7 out of 20 terms for men, compared to only one out of 20 terms for women).

In general, we can observe that Twitter speech appears to be more personalized and emotional, while Knesset floor utterances appear to be more formalistic and procedural. Unsurprisingly, Twitter speech is distinguished by the use of emojis (albeit mostly by female posters).

| Male | Female |
|---|---|
| אמר (said) | ילד (child) |
| יושב (chair/sitting) | איש (person) |
| ראש (head) | אמת (truth) |
| אדון (mister) | מיני (sexual) |
| תורה (Torah) | שמחה (happy/happiness) |
| עמית (friend/colleague) | תודה (thanks) |
| ערבי (Arab) | הורה (parent) |
| הכנסה (income/insertion) | חיים (life) |
| צה״ל (IDF) | אלימות (violence) |
| חבר (friend) | מוגבלות (disability) |
| בטוח (safe/certain) | חברה (society/company) |
| ביטחון (security) | משפחה (family) |
| רשב״י (Rashbi) | טיפול (treatment) |
| יהודי (Jewish) | נפשי (psychological) |
| חרדי (Haredi) | נפגע (victim) |
| יישוב (settlement) | אזרחיות (female citizens) |
| שמח (happy) | מקלט (shelter) |
| מכובד (esteemed) | פסולת (garbage) |
| מילואים (reserve duty) | זכות (right) |
| כנסת (Knesset) | בת (daughter) |

Table 4: Top 20 Distinguishing Terms – Knesset (Ranked Top to Bottom)

| Male | Female |
|---|---|
| ישראל (Israel) | איש (person) |
| ממשלה (government) | 💜 |
| בנט (Bennett) | רוסי (Russian) |
| לפיד (Lapid) | מיני (sexual) |
| יהודי (Jewish) | חברה (society/company) |
| פטריוט (patriotic) | גבר (man) |
| מדינה (state) | שמחה (happiness) |
| ליכוד (Likud) | 👩 |
| צה״ל (IDF) | תודה (thanks) |
| טרור (terror) | ילד (child) |
| שר (minister) | אמא (mother) |
| נתניהו (Netanyahu) | נהדר (great) |
| עוצמה (power) | ריגש (emotionally moving) |
| בג״ץ (supreme court) | 😊 |
| ינוניוז (YINONEWS) | עאידה (Aida) |
| מחבל (terrorist) | דיון (discussion) |
| כחול (blue) | אהובה (female loved one) |
| ביטחון (security) | 👻 |
| גנץ (Gantz) | שוויון (equality) |
| ראש (head) | חמוד (cute) |

Table 5: Top 20 Distinguishing Terms – Twitter (Ranked Top to Bottom)

## 6. Conclusion

We present IsraParlTweet, a new corpus of Knesset discussions (between the years 1992-2023) and Twitter posts (between the years 2008-2023) in the Hebrew language, linked by speaker and to additional metadata. In addition to raw text, the corpus contains rich meta-data in the form of speaker & session details, as well as morphological and sentiment tagging. The parallel nature of the corpus, along with the rich meta-data, provide a unique opportunity to study both the formal (Knesset) and the informal (Twitter) discourse in the Israeli political arena and explore the connections and interactions between them. We demonstrate this potential by utilizing a single speaker attribute – gender – to perform extensive statistical analyses which yield intriguing results and observations.

IsraParlTweet is one of the largest resources in spoken Hebrew, and the first to make the Knesset plenum debate and MKs social media messages accessible to researchers. It is our hope that this resource will facilitate new and exciting studies in a variety of acedemic research fields.

## 7. Acknowledgements

This research was supported by the Israel Science Foundation (Grant No. 2315/18). The authors would like to thank Noa Amir, Uri Mishmar, Hagar Kaminer, Batel Yaakov and Roni Shapira for diligent research assistance; Gili Goldin and Lior Schwartz for providing the metadata; and finally, deep gratitude is extended to Knesset Head of BI Shoshana Makover for significant assistance and guidance in procuring Knesset data.

## 8. Bibliographical References